\documentclass[runningheads]{llncs}

\usepackage[T1]{fontenc}
\usepackage{graphicx}
\usepackage{amsmath}
\usepackage{subcaption}
\usepackage{algorithm}
\usepackage{algpseudocode}
\usepackage{booktabs}
\usepackage{appendix}
\begin{document}
\title{Joint 2D-3D Segmentation and Association in Street-level Imaging
}

\titlerunning{Joint 2D-3D Segmentation and Association}
\author{Amir Melnikov  \and
Masayuki Tanaka \and
Yusuke Monno \and
Masatoshi Okutomi
}
\authorrunning{A. Melnikov et al.}
\institute{
Institute of Science Tokyo, 2-12-1 Ookayama, Meguro-ku, Tokyo, Japan 152-8550 \\ \email{amelnikov@vip.sc.eng.isct.ac.jp} \vspace{3mm}
\\ Project page: \url{http://www.ok.sc.e.titech.ac.jp/res/Seg2D3D/}
}
\maketitle
\begin{abstract}
Accurate interpretation of street-level imagery is essential for large-scale urban mapping and the creation of Spatial Digital Twin (SDT) environments. This work presents a unified framework for joint 2D–3D segmentation and association that integrates visual semantics with multi-view geometric reasoning. Unlike conventional approaches that rely heavily on sequential frames for temporal tracking, our method leverages zero-shot detection and segmentation together with structure-from-motion reconstruction to establish stable cross-view correspondences. A 3D-driven association mechanism replaces traditional 2D multi-object tracking, using geometric consistency to guide identity preservation across wide-baseline viewpoints and varying imaging conditions. By combining 2D texture cues with global 3D context, the proposed pipeline is well-suited for scalable street-level processing and can be used for a variety of object types. Experiments demonstrate substantially improved coverage of ground-truth sequences and more robust identity retention compared to state-of-the-art 2D-only tracking methods, achieving a \textbf{22\%} performance gain in challenging urban scenarios.
\end{abstract}


\begin{figure}[h]
    \centering
    \includegraphics[width=\textwidth]{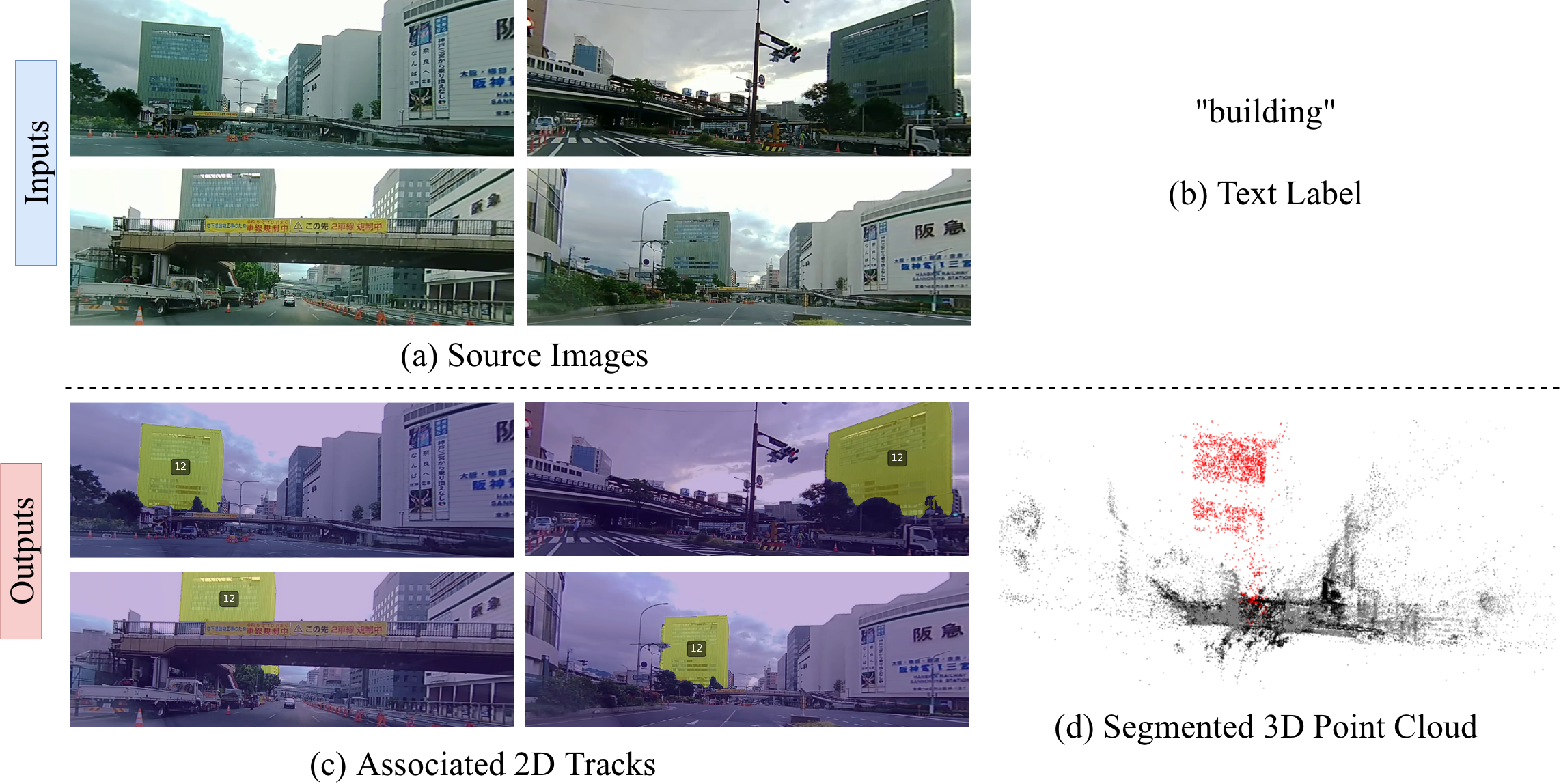}
    \caption{Overview of the proposed pipeline's inputs and outputs. Multi-view 2D images are used to generate the 3D model, which is then used to correlate the keypoints to associate segments to the same real-world 3D objects. Additionally, a segmented 3D point cloud is generated, seen in (d). In red are the 3D Points associated with building ID 12.}
    \label{fig:blackbox}
\end{figure}

\section{Introduction}

Multi-view street-level imagery is a crucial component in constructing reliable photorealistic virtual representations of static environments, commonly referred to as \emph{Spatial Digital Twin} (SDT) models \cite{Shiode2000Urban}. These models form the foundation for a range of applications, from urban planning and navigation to simulation, virtual reality, and autonomous systems. Previous works have established fundamental datasets and pipelines for training perception and navigation models that utilize SDT data combined with real-world imagery, enabling context-aware visual understanding of physical spaces \cite{Geiger2012KITTI}. The fidelity of such systems relies not only on the accuracy of geometric modeling but also on the ability to understand and classify the
abstract objects
or structures represented visually within the data.

Data in SDT environments can generally be divided into two complementary domains: \emph{geometric} or \emph{volumetric data} and \emph{texture (visual) data}. Geometric data, typically captured through LiDAR scanning or photogrammetric reconstruction, provides accurate spatial and structural representations of the scene~\cite{Seitz2006MVS}. This kind of data can be visualized and used as a point cloud, mesh models, block models, or similar representations of the 3D spatial information. Texture data, on the other hand, is derived from RGB or multispectral imagery captured from various viewpoints, contributing the appearance and semantic context necessary for realistic rendering and interpretation. Bridging these two domains remains a central challenge for achieving consistent and semantically meaningful information across modalities.

In this work, we propose a joint 2D-3D framework that combines texture-level 2D semantic segmentation with 3D association tracking to bridge the gap between visual and geometric domains. By establishing consistent object identities across 2D frames and linking them to 3D reconstructions, our approach utilizes 3D information to contextualize texture information, combining the two modalities in a way suitable for large-scale SDT environments, with less reliance on sequential data collection for temporal tracking methods. Specifically, our suggested pipeline takes the unsorted source images, along with a text label of the desired objects, and produces as output 3D segmented point clouds as well as 2D association tracks of 2D segments, as can be seen in Figure \ref{fig:blackbox}.

To further enhance spatial consistency, the pipeline integrates a 3D-based substitute for what would be multi-object tracking (MOT) techniques, replacing traditional 2D keypoint-based tracking methods. Conventional feature tracking, such as optical flow or descriptor matching, often suffers from drift and instability in complex scenes due to repetitive structures, occlusions, and perspective changes \cite{Lucas1981Optical}. Most importantly, they rely heavily on sequential data and perform much worse on non-sequential images with changing perspectives as is usually acquired in a large-scale static scene. Utilizing the 3D-aware features constructed in the SfM models \cite{Schonberger2016SfM} allows assigning all 2D segments of the same object to the same 3D points, achieving a robust and grounded result.

\section{Related Work}

\subsubsection{2D--3D Detection, Segmentation, and Multi-Object Tracking}

Associating 2D image regions with 3D scene structure has been significantly advanced by modern vision foundation models. Zero-shot object detection, such as \emph{GroundingDINO} \cite{Liu2024GroundingDINO}, leverages transformer-based architectures and text-aligned query generation to localize semantically meaningful regions without dataset-specific training. This is especially relevant for urban scenes, where annotated datasets are limited and building-related categories are often absent from standard benchmarks.

For segmentation, the \emph{Segment Anything Model} (SAM) \cite{Radford2021CLIP,Kirillov2023SAM} introduced a prompt-driven framework capable of producing high-resolution masks across diverse environments. Although SAM does not assign semantic labels, it provides accurate pixel-level boundaries necessary to delineate facade regions and architectural components.

These capabilities are combined in \emph{Grounded-SAM} \cite{Ren2024GroundedSAM}, which integrates GroundingDINO’s text-conditioned detections with SAM’s segmentation masks. Grounded-SAM produces semantically meaningful instance masks in a zero-shot manner, making it particularly suitable for facade extraction in large-scale urban datasets. When paired with SfM outputs such as COLMAP \cite{Schonberger2016SfM}, these 2D instances can be projected into 3D, enabling cross-view semantic consistency checks and association estimation.

Temporal consistency across frames is commonly approached using detection-based multi-object tracking (MOT). Detection-based MOT maintains persistent object identities by associating detections across frames \cite{Bewley2016SORT,Wojke2017DeepSORT}. However, temporal-focused methods cannot be used with datasets with non-sequential images—the kind COLMAP is most useful for when generating a 3D scene. Therefore, feature-based tracking and re-ID methods are used. While re-ID itself is mostly limited to human tracking \cite{Zheng2015Market1501}, feature-based transformer MOT, such as MOTRv2 \cite{Zhang2023MOTRv2}, is able to perform to some extent even with non-sequential data.

\subsubsection{Semantic segmentation in 3D}
Semantic segmentation of 3D objects from street-level imaging can be approached in several ways. One strategy involves the \emph{propagation of semantic labels on video data}~\cite{ref1}, which requires sequential frames and employs feature-based tracking to maintain temporal consistency. However, this method is computationally intensive and often uses frame skipping at the cost of reduced completeness. Alternative methods perform \emph{class-unaware associations in 3D point clouds}~\cite{ref2} or use \emph{super point graphs (SPG)}~\cite{ref3}, but these approaches neglect the higher precision achievable through 2D semantic segmentation, as they rely solely on feature embeddings within the point cloud.

More recent research~\cite{ref4} explores joint 2D-3D networks leveraging visual and geometric data. However, rather than relying on indoor RGB-D voxelized representations, which limit scalability, our novelty lies in replacing traditional temporal tracking with a geometric, 3D-driven association tailored for complex outdoor urban settings.

\section{Methodology}
\begin{figure}[t]
    \centering
    \includegraphics[width=\textwidth]{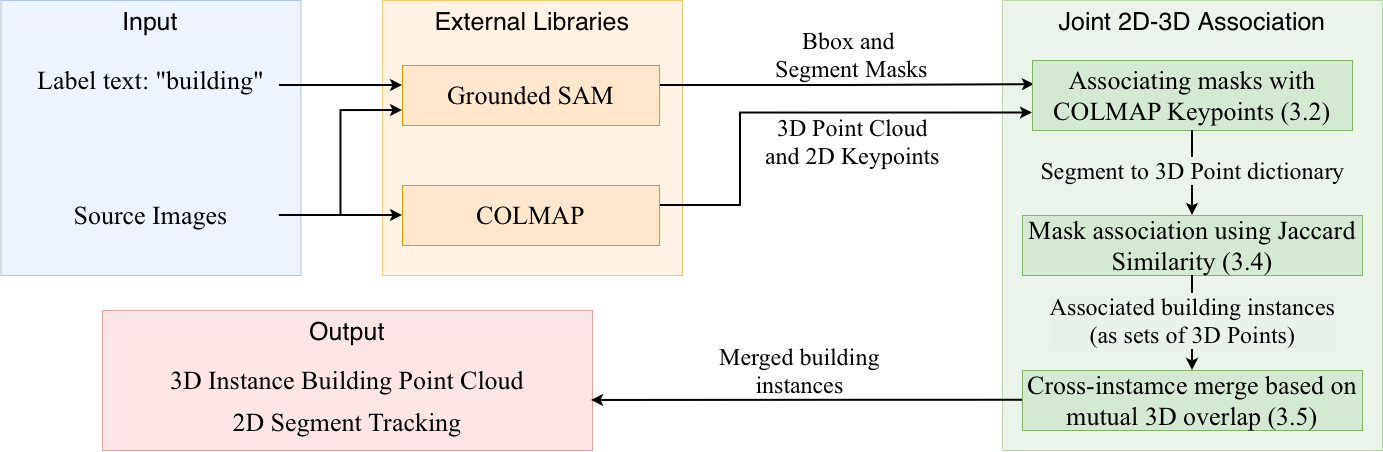}
    \caption{Overview of the proposed multi-stage pipeline. The input images are first processed with Grounded SAM to generate detections and segmentation masks. COLMAP keypoints are projected onto the masks, and their associated 3D point tracks are used to identify persistent correspondences across views. The associated mask sets are then clustered into building-level instances based on shared 3D point associations. (Corresponding subsection numbers are in parenthesis).}
    \label{fig:overall_process}
\end{figure}

\begin{figure}[t]
    \centering
    \includegraphics[width=\textwidth]{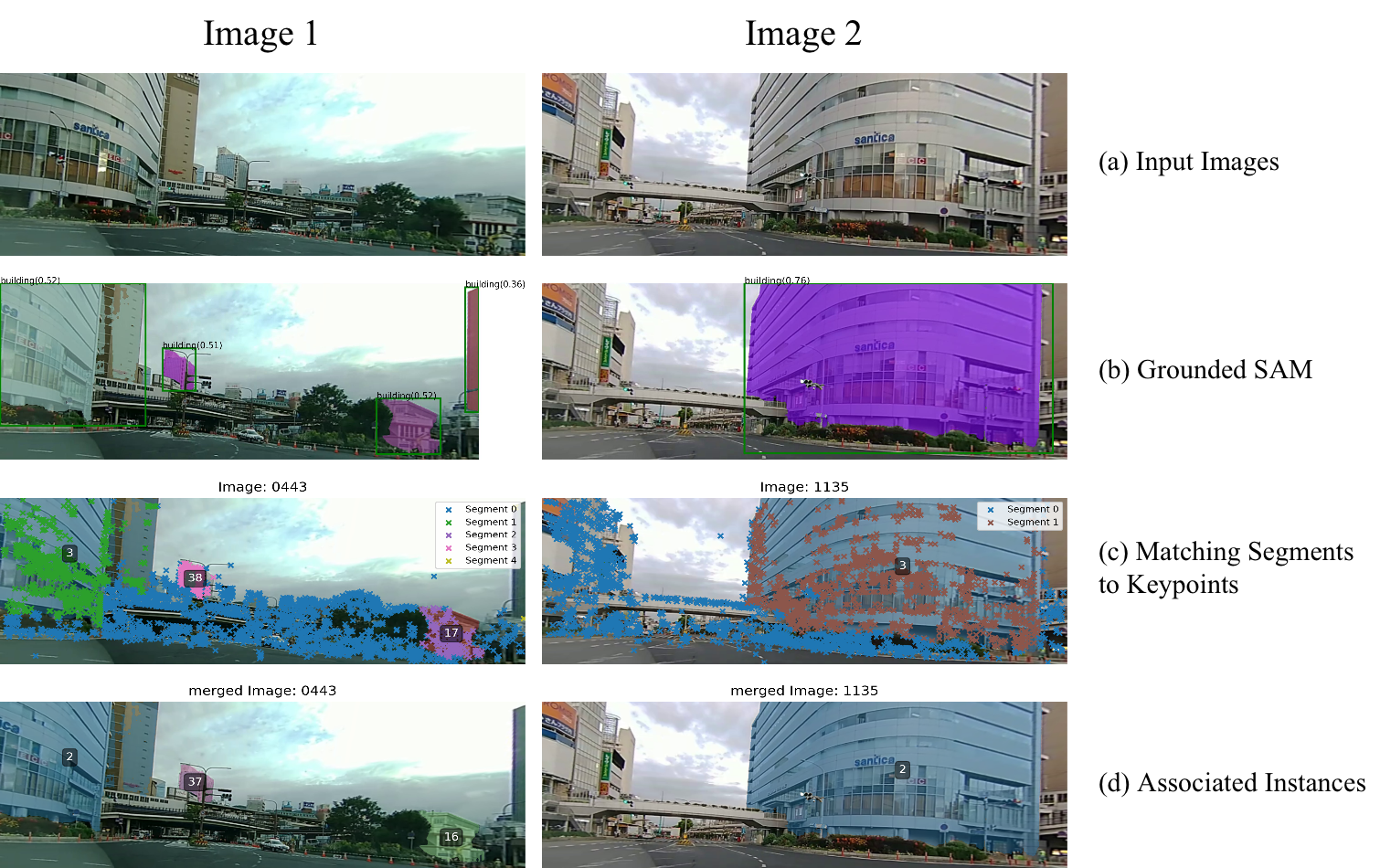}
    \caption{Visualization of multiple perspectives of the 3D object with the key stages in the proposed processing pipeline. 
(a) Input source image. 
(b) Grounded SAM output. 
(c) COLMAP 2D keypoints overlaid on the masks and linked to their corresponding 3D point IDs via track associations. 
(d) associated mask sets forming complete building instances by grouping segments that share common 3D points and merging building instances (building~3 in (c) merged into building~2 in (d)).
}
    \label{fig:2x4_grid}
\end{figure}

The proposed pipeline constructs consistent building-level correspondences by combining zero-shot 2D segmentation with multi-view geometric reconstruction from COLMAP. Given a set of input images $\mathcal{I} = \{I_1, \dots, I_N\}$, the objective is to produce a set of building instances, each represented by the collection of its corresponding 2D masks across images (seen in Figure \ref{fig:blackbox}(c)) and the associated set of 3D points belonging to that object (seen in Figure \ref{fig:blackbox}(d)). The pipeline proceeds in five stages: (1) zero-shot detection and segmentation, (2) association of segmentation masks with COLMAP keypoints, (3) mapping these keypoints to 3D points via COLMAP tracks, (4) association of masks into facade-level groups using 3D Jaccard similarity, and (5) merging building groups that exhibit sufficiently large mutual 3D point overlap. The overall process is shown in Figure~\ref{fig:overall_process}, and the main stages are visualized in Figure \ref{fig:2x4_grid}. 

\subsection{Zero-shot detection and segmentation}

Each image $I_i$ is processed with Grounded-SAM, which fuses the text-conditioned detection capabilities of GroundingDINO with the high-resolution segmentation masks produced by SAM. For each image, a set of instance masks is extracted,
\[
\mathcal{S}_i = \{ S_{i,1}, \dots, S_{i,m_i} \},
\]
where each mask $S_{i,k}$ is a binary region in the image plane.

\subsection{Associating masks with COLMAP keypoints}

For each image, COLMAP provides a set of detected 2D keypoints, stored in the standard format:
\[
[\mathit{Point2D}_\mathit{ID
},\, x,\, y,\,\mathit{Point3D}_\mathit{ID}]
\]
where each keypoint has pixel coordinates $(x,y)$ and, when available, a reference to the ID of the corresponding reconstructed 3D point. These keypoints exist in the same pixel coordinate frame used by Grounded-SAM, allowing direct spatial comparison. To associate each segmentation mask with its underlying geometric observations, every keypoint is tested for membership inside the mask region. If a keypoint lies within the binary mask, it becomes part of that segment’s observation set. In this way, each segmented region inherits the set of COLMAP features that physically fall inside it, and the mask later receives a segment identifier that is appended to the keypoint entry.

\subsection{Mapping keypoints to COLMAP 3D points}

COLMAP represents each reconstructed 3D point using a compact record of the form

\[
[\texttt{ID},\, X,\, Y,\, Z,\,  R, G, B, \varepsilon,\, 
\texttt{TRACK}]
\]
where:
\[
\mathbf{P}_{\mathrm{3D}} = 
\left\{
\begin{aligned}
&\mathrm{ID}        &&\text{: unique 3D point identifier},\\
&X,Y,Z              &&\text{: reconstructed 3D coordinates},\\
&R,G,B              &&\text{: estimated RGB color},\\
&\varepsilon        &&\text{: reprojection error},\\
&\texttt{TRACK}     &&\text{: list of } [\mathit{Image}_\mathit{ID},\; \mathit{Point2D}_\mathit{ID
}]
\end{aligned}
\right.
\]

\begin{figure}[t]
    \centering
    \includegraphics[width=\textwidth]{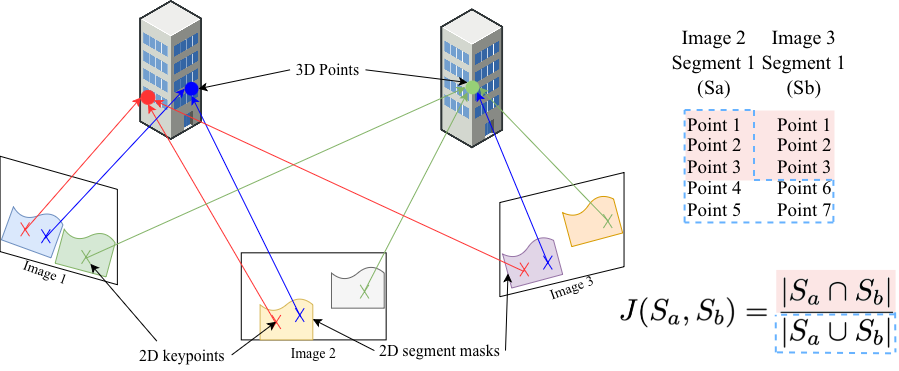}
    \caption{Relationship between 2D COLMAP keypoints, Grounded-SAM generated segment masks and 3D points. In the example, even though the blue segment may have 3D points in other objects, the majority will point towards the correct object.}
    \label{fig:3d_relationship}
\end{figure}

The \texttt{TRACK} field is comprised of a list of $[\mathit{Image}_\mathit{ID},\; \mathit{Point2D}_\mathit{ID
}]$, meaning it contains all 2D observations that contributed to this reconstruction. Each track entry is a tuple that refers back to a specific image and a specific 2D keypoint. Once the mask association is known, recovering 3D points belonging to a segment is straightforward: any 3D point whose track includes a keypoint assigned to the mask is considered part of the 3D support of that segment. A visualization of this relationship can be seen in Figure \ref{fig:3d_relationship}. 

This step effectively lifts the segmentation from 2D into 3D, using COLMAP’s multi-view correspondence to identify which reconstructed points are geometrically supported by each mask. Since tracks encode the full multi-view history of every 3D point, the resulting association is viewpoint-independent and robust to appearance changes across images. The segment therefore becomes not only a 2D region but is transformed into a set $S_a$ containing all relevant 3D point IDs associated with this 2D segment.

\subsection{Mask Association using Jaccard similarity}

The basic working logic to connect the 2D masks is that masks belonging to the same building should share many of the same 3D points across viewpoints, since they are observing the same real-life 3D object. To quantify this relationship, a Jaccard similarity \cite{Jaccard1901} between two sets is defined for any two sets $S_a$ and $S_b$ as
\begin{equation}
J(S_a, S_b)
=
\frac{
    \lvert S_a \cap S_b \rvert
}{
    \lvert S_a \cup S_b \rvert
},
\end{equation}
where the sets $S_i$ are the sets containing a list of the 3D point IDs associated with a 2D segment mask, as explained in subsection 3.3.

Association begins with the mask having the largest number of associated 3D points, which is designated as the initial seed. All remaining masks whose 3D Jaccard similarity with the current group exceeds a threshold $\tau_J$ are merged into the building instance. The group’s set of 3D points that is used for the comparison is updated after each merge, strengthening the geometric signature. This allows instances to grow disproportionately when masks are not consistent, but it has not shown a problem in this case. Since any 3D point can be linked to multiple 2D keypoints, the same 3D point can be associated eventually with multiple instances. An example can be seen in Figure \ref{fig:3d_relationship}: the green 3D point is associated with the wrong segment in Image 3. However, given enough 2D keypoints and 3D points associated to this segment, the association will be to the correct 3D object.

\subsection{Cross-instance merging based on mutual 3D overlap}

Following the first association stage, some buildings may remain artificially separated across different initial clusters, often due to the different 2D keypoints sometimes assigning their linked 3D point to multiple building instances. To resolve these cases, the pipeline performs a second-level merging procedure at the building-instance level. For any two building instances $B_u$ and $B_v$ with associated 3D point sets $P_{B_u}$ and $P_{B_v}$, their mutual 3D overlap ratio is defined as

\begin{equation}
J(B_u, B_v)
=
\frac{
    \lvert P_{B_u} \cap P_{B_v} \rvert
}{
    \lvert P_{B_u} \cup P_{B_v} \rvert
}.
\end{equation}

A merge is triggered when
\begin{equation}
J(B_u, B_v) \ge \tau_M ,
\end{equation}
where $\tau_M$ is a threshold that determines how much 3D point sharing is required for two building groups to be considered identical. When merging occurs, the union of 3D points and all masks belonging to both instances is combined into a single building identity (instance). This ensures that facades captured across non-overlapping images are still unified when their 3D structure agrees. This can be seen in Figure \ref{fig:2x4_grid}: building no. 3 seen in (c) is merged into building no. 2 generated from other views in (d). 

At this step, instances with a small number of 3D points \( n \le n_{\min} = 10 \) were discarded. Finally, we create a reverse lookup dictionary to decide for each 3D point a single instance it belongs to based on the majority of the associated segments.
A summary of this process can be seen as pseudo-code in Algorithm~\ref{alg:association}. 
\subsection{Parameter ranges}

\begin{algorithm}[t]
\caption{Two-stage association of building instances using 3D Jaccard and cross-instance merging}
\label{alg:association}
\begin{algorithmic}[1]
\Require Masks $S_{i,k}$ with 3D point sets $\mathrm{pt}(S_{i,k})$; thresholds $\tau_J$ and $\tau_M$; minimum point threshold $n_{\mathrm{min}}$.
\State Perform initial mask-level association using 3D Jaccard similarity (as described in Section 3.4), producing instances $\mathcal{B} = \{B_1,\dots,B_m\}$ with associated point sets $P_{B_1},\dots,P_{B_m}$.

\Repeat
    \State Set $\text{changed} \leftarrow \text{false}$.
    \For{each unordered pair $(B_u, B_v)$}
        \State Compute overlap 
            \[
            \mathrm{J}_{}(B_u, B_v)
            =
            \frac{
                |P_{B_u} \cap P_{B_v}|
            }{
                |P_{B_u} \cup P_{B_v}|
            }
            \]

        \If{$J \ge \tau_M$}
            \State Merge $B_u$ and $B_v$ into a new instance $B^\prime$.
            \State Update its 3D point set $P_{B^\prime} = P_{B_u} \cup P_{B_v}$.
            \State Replace $B_u$ and $B_v$ with $B^\prime$ in $\mathcal{B}$.
            \State Set $\text{changed} \leftarrow \text{true}$.
            \State \textbf{break}
        \EndIf
    \EndFor
\Until{$\text{changed} = \text{false}$}

\State \Return Final merged $\mathcal{B}$.
\end{algorithmic}
\end{algorithm}

The framework relies on a small set of thresholds and confidence values governing mask filtering, 3D point aggregation, and text/detection quality.
The chosen ranges were obtained empirically and were found to generalize well across the tested datasets. The values for text and detection confidence are the recommended confidence scores for Grounded-SAM. 
A complete list of recommended parameter values is provided in Appendix A.

\section{Experiment}
\subsection{Datasets}
The main dataset \emph{Dataset 1} used for analysis in this work was collected by MICWARE Inc. in the intersection outside Sannomiya Station in Kobe, Hyogo Prefecture, Japan, a dataset that was used in our previous works \cite{MonnoMicware2025}.

The second dataset used for comparison purposes is CityScapes \cite{Cordts2016Cityscapes}, specifically the training set captured in Zurich. This dataset consists of 3660 images captured in 121 separate sequences, driving through various locations in the city. This dataset is not optimized for 3D model generation due to the low reappearance of the buildings in the separate sequences. COLMAP was used with exhaustive matching to try to combine as many image features as possible. Therefore, only a subset of the images consisting of 115 images, matched by COLMAP to generate a single model was used. 

\subsection{Ground Truth Annotation}

For quantitative comparison of the ability to match the 2D segments of building facades, ground truth data was generated for \emph{Dataset 1}. Using \emph{Computer Vision Annotation Tool} (CVAT) \cite{CVAT2025}, ground truth bounding boxes were marked manually by a human for 30 separate buildings in the scene. The buildings had consistent ID throughout all frames in the dataset, resulting in multiple tracks per building. Overall, for the 8-turn subset, 1503 ground truth bounding boxes were generated, out of which 24 were auto-interpolated using CVAT interpolation and were verified by the annotator.
Due to the labor-intensive aspect of manual annotation, only \emph{Dataset 1} was manually annotated and thus is able to view comparative quantitative results. The same processing methods were used on the other datasets for qualitative comparison. 

\subsection{Implementation Details}

The proposed pipeline integrates three core components: Grounded-SAM for zero-shot segmentation, COLMAP for multi-view geometric reconstruction, and a 3D point–based instance association module. Grounded-SAM was executed using GroundingDINO-T + SAM Vit-H with the text label \texttt{building}. COLMAP was run using exhaustive matching for CityScapes and for the MICWARE datasets, followed by standard incremental reconstruction. Keypoints were extracted using COLMAP’s default SIFT implementation. All experiments were run on a workstation with an NVIDIA RTX4080 GPU and 64GB RAM.

\subsection{Comparison Methods}

Two comparison baselines were evaluated against the proposed approach.  
The first is a naive geometric baseline in which segments are associated strictly using per-frame 2D Intersection-over-Union (IoU) with the previous frame. 
The second baseline is a state-of-the-art 2D video segmentation tracker combining SAM2 with MOTRv2, a more modern implementation of VISAM. SAM2 generates per-frame segmentation masks and MOTRv2 links them into temporal tracks via end-to-end transformer-based instance association. These methods represent strong 2D-only baselines for object-level tracking in traditional video tracking challenges.

\section{Evaluation}
\subsection{Evaluation Metrics}
The evaluation focuses on the ability of each method to produce consistent building-level instance association across multi-view imagery. Because the problem is fundamentally different from classical object tracking, conventional MOT metrics \cite{Luiten2021HOTA} such as IDF1, MOTA, and HOTA are not suitable for our task.
These metrics heavily penalize instances going in and out of frame, even with correct ID assignment. These metrics depend on object persistence across adjacent frames, assume dense temporal continuity, and are meant to evaluate the entire system, including detection and segmentation performance (which are external to this work). Instead, we introduce metrics based on proper instance association capability, along with qualitative visual comparisons. For reference, traditional MOT metrics can be found in Appendix B.

In contrast, building facades under multi-view reconstruction reappear intermittently across disparate viewpoints and non-sequential frames and must be associated through geometry rather than motion. Therefore, these metrics systematically under-represent methods that rely on 3D consistency and are not suitable for the task addressed here.

To assess cross-view instance consistency, we introduce two metrics that more truthfully represent the ability to correctly assign texture data to the 3D space: \emph{Coverage} and \emph{Adjusted Coverage}.

\subsubsection{Coverage}

Coverage measures how consistently a tracker retrieves the correct building identity across all frames where that building appears in. It is defined as:

\begin{equation}
\text{Coverage}
=
\frac{
    \#\,\text{GT frames correctly matched}
}{
    \#\,\text{Total GT frames of the instance}
}.
\end{equation}

Coverage reflects the tracker's ability to keep a consistent identity, but it is sensitive to frames where the detection system fails to produce a segment at all. Such failures are not the result of the tracking algorithm, but of the upstream segmentation or detection stage. To account for these missing detections, we define \emph{adjusted coverage}.

\subsubsection{Adjusted Coverage}
Let:
- $\#\text{(MissedSeg)}$ be the number of GT frames where the segmentation or detection stage produced no usable mask (i.e., the tracker never had a chance to match the object).

Then the adjusted coverage is:

\begin{equation}
    \text{Adjusted Coverage} =
    \frac{
    \#\text{(GT frames correctly matched)}
    }{
    \#\text{(Total GT frames)} - \#\text{(MissedSeg)}
    }.
\end{equation}

This adjustment isolates the performance of the \emph{association} by removing frames where the failure originates from missing or incorrect initial detections, ensuring the metric reflects identity consistency rather than upstream segmentation errors.

A perfect tracker yields:
\begin{equation}
\text{Coverage} = \text{Adjusted Coverage} = 1.
\end{equation}
Fragmentation reduces adjusted coverage, because even if individual predicted segments perform well locally, they jointly cover fewer of the required ground-truth frames. Furthermore, in the case of multi-sequence datasets, if a tracker is not designed to operate across multiple sequences (or is not sequence-agnostic), its coverage score is upper-bounded by the proportion of the longest sequence with respect to the entire dataset.

\subsection{Quantitative Results}
\begin{figure}[t]
    \centering
    \includegraphics[width=\textwidth]{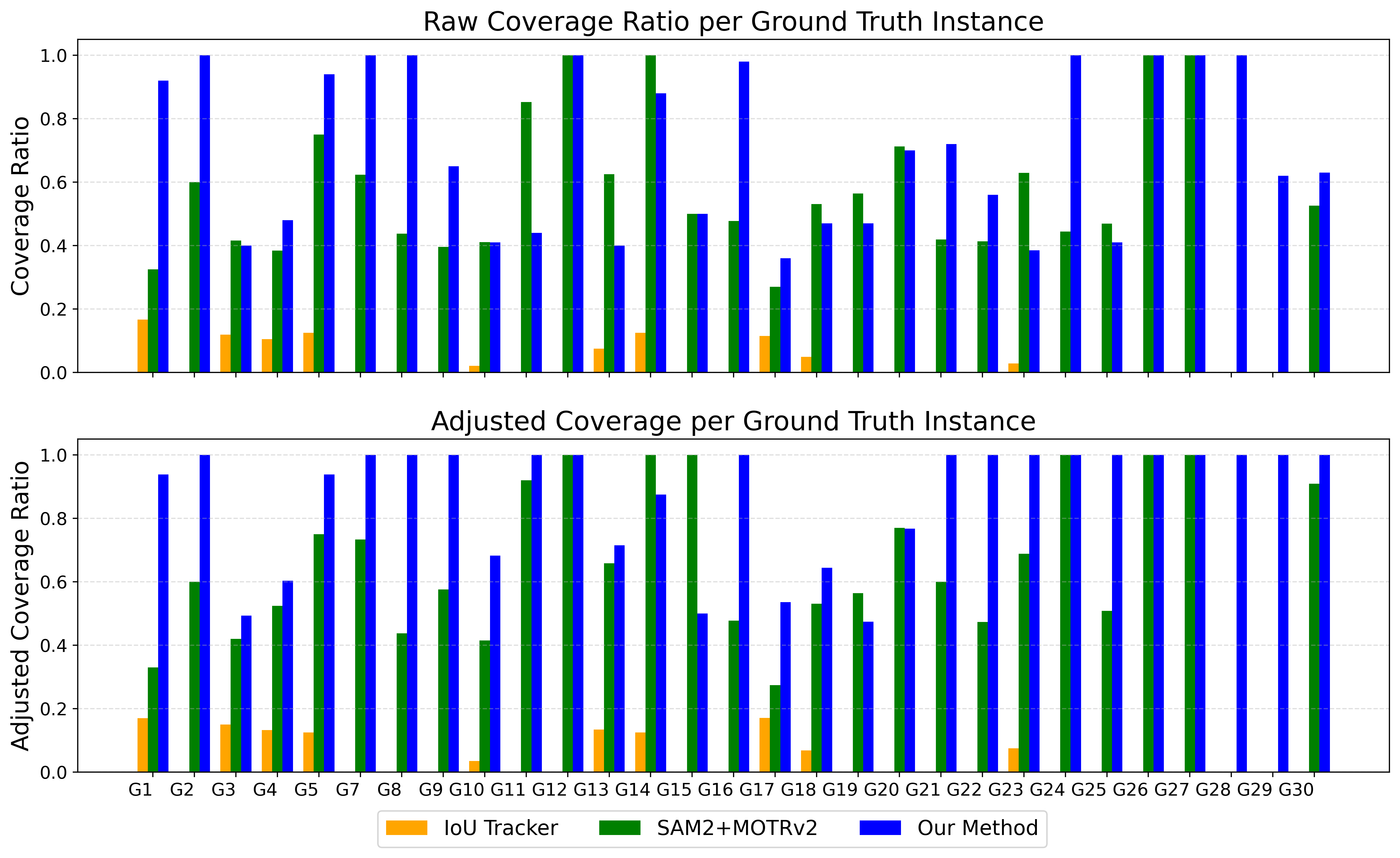}
    \caption{Coverage and Adjusted Coverage results for the three methods, per Ground Truth Instance across Dataset 1. Higher is better.}
    \label{fig:coverage_combined}
\end{figure}

We evaluate three tracking and association methods—IoU-based tracking, SAM2 with MOTRv2, and our proposed 3D-guided approach—using the Coverage and Adjusted Coverage metrics across all annotated buildings in \emph{Dataset 1} (Fig.~\ref{fig:coverage_combined}, Tab.~\ref{tab:quantitativesummary}). The IoU-based tracker performs the poorest, with very low Coverage (0.038) and Adjusted Coverage (0.051), indicating frequent identity failures even under minor viewpoint changes. SAM2+MOTRv2 achieves moderate performance, with Coverage of 0.533 and a higher Adjusted Coverage of 0.606, suggesting that part of its errors stem from segmentation inconsistencies rather than tracking. The proposed 3D-guided method yields the best results, with Coverage of 0.655 and Adjusted Coverage of 0.841—a substantial improvement over the baselines—showing that enforcing multi-view geometric consistency significantly enhances identity preservation. Notably, considering the evaluated subset consists of 8 sequences, both SAM2+MOTRv2 and our proposed approach exceed the theoretical upper bound of 0.125 that would occur if multi-sequence association was not handled, further confirming their ability to maintain consistent identities across multiple sequences.

\begin{table}[t]
\centering
\caption{Average Coverage and Adjusted Coverage for the evaluated methods.}
\vspace{2mm}
\label{tab:quantitativesummary}
\begin{tabular}{lcc}
\toprule
\textbf{Method} & \textbf{Coverage ~} & \textbf{Adjusted Coverage} \\
\midrule
IoU Tracker & 0.038 & 0.051 \\
SAM2+MOTRv2 & 0.533 & 0.606 \\
\textbf{Our Method} & \textbf{0.655} & \textbf{0.841} \\
\bottomrule
\end{tabular}
\end{table}

\subsection{Qualitative Results}

Examples of the associated building instance models and per-frame mask assignments are visualized for all datasets. The proposed approach shows consistent instance identity across distant viewpoints, while the 2D-only baselines often fragment or merge facades incorrectly. Visualizations of typical errors in tracking are shown in Figure~\ref{fig:errors_vis}. Additional videos showing the complete tracking and association result in 2D, as well as some sample instance 3D models visualized in Figure \ref{fig:blackbox} with CloudCompare \cite{CloudCompare}, are available at the Project Page: \url{http://www.ok.sc.e.titech.ac.jp/res/Seg2D3D/}

\begin{figure}[t]
    \centering
    \includegraphics[width=\textwidth]{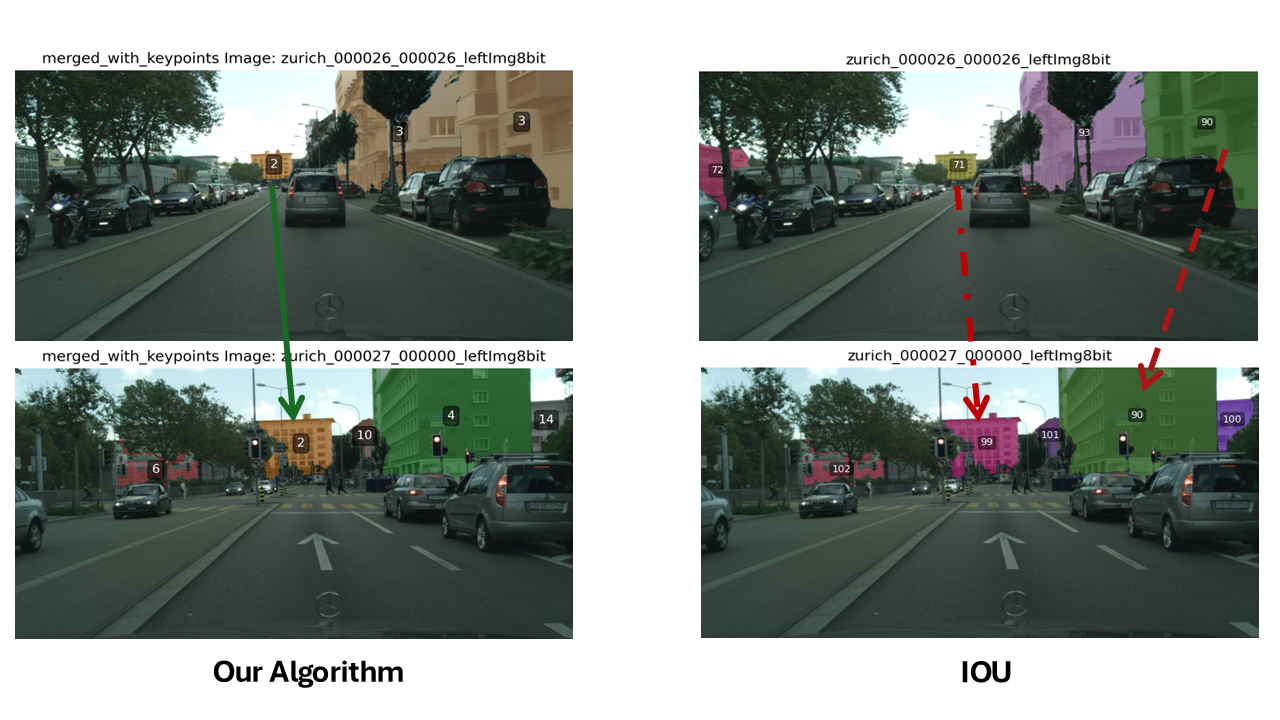}\\
    \vspace{-3mm}
    \caption{Typical errors generated with motion tracking compared to our method. The two frames (top and bottom) are from separate sequences in \emph{Dataset 2}. Our method correctly assigns ID 2 to the building in the center of the frame (marked with a green arrow). IoU is wrong both in the ID assignment for this building (dashed-dotted red arrow) and incorrectly assigns the same ID to two separate buildings at the edge of the frame, inducing ID spillover (dashed red arrow).}
    \label{fig:errors_vis}
\end{figure}

\section{Discussion and Future Work}

The results presented in this work demonstrate the benefits of incorporating multi-view geometric consistency for persistent facade 2D-3D association in complex urban scenes. While this method proved effective and was able to surpass MOT trackers at this task, some limitations remain. The pipeline depends strongly on the quality of external detection and segmentation models. As indicated by the gap between raw Coverage and Adjusted Coverage, upstream detection failures directly impact the overall performance, independent of the association strategy. Furthermore, full SfM reconstruction introduces computational complexity compared to 2D tracking, limiting scalability for massive datasets and real-time use. Future work could mitigate this via hierarchical SfM or chunking datasets. Additionally, while our empirically derived Jaccard thresholds (Appendix A) suit large, static structures like buildings, scaling to complex scenes with smaller or dynamic object classes will likely require adaptive, learning-based parameter estimation. Finally, extending the framework with incremental SfM, online segmentation, and dataset-specific fine-tuning could reduce missed detections and support live digital-twin updates or continuous monitoring.

\subsubsection{Acknowledgements} This research was conducted as part of a collaborative research project with MICWARE CO., LTD.

\newpage
\appendix
\section*{Appendix A: Parameter Values}

Table~\ref{tab:params2} details the parameters that were used both with external libraries and as thresholds in our algorithm, as explained in Section 3.6.
\begin{table}[h!]
\vspace{-2em}
\centering
\caption{Recommended parameter values for the pipeline.}
\vspace{2mm}
\begin{tabular}{ll}
\toprule
Parameter & Range \\
\midrule
Jaccard mask threshold $\tau_J$ & $0.15$--$0.30$ \\
Building overlap threshold $\tau_M$ & $0.10$--$0.25$ \\
Minimum 3D points per instance $n_{\mathrm{min}}$ & $5$--$20$ \\
Detection confidence & $0.2$--$0.5$ \\
Text confidence & $0.2$--$0.4$ \\
\bottomrule
\end{tabular}
\label{tab:params2}
\vspace{-0.9em}
\end{table}

\vspace{-0.8em}
\section*{Appendix B: Additional Metrics Result}

Table~\ref{tab:motmetrics} details additional traditional MOT metrics results for all three tracking methods as explained in Section 5.1.

\begin{table}[h!]
\vspace{-0.9em}
\centering
\caption{Traditional MOT metrics comparison across the three methods.}
\label{tab:motmetrics}

\begin{subtable}{\columnwidth}
\centering
\caption{Identity-based metrics}
\small
\begin{tabular}{lccccccc}
\toprule
Method & Frames & IDF1 & IDP & IDR & IDs & FM \\
\midrule
MyAlgo         & 1763 & 1.1\%  & 1.2\%  & 1.1\%  & 2  & 4 \\
IOUtracker     & 1763 & 2.8\%  & 2.7\%  & 2.9\%  & 12 & 61 \\
SAM2+MOTRv2    & 1763 & 21.8\% & 14.2\% & 47.1\% & 63 & 19 \\
\bottomrule
\end{tabular}
\end{subtable}

\vspace{-0.7em}

\begin{subtable}{\columnwidth}
\centering
\vspace{3mm}
\caption{Detection, trajectory, and overall metrics}
\small
\begin{tabular}{lccccccccc}
\toprule
Method & Rcl & Prc & MT & PT & ML & FP & FN & MOTA & MOTP \\
\midrule
MyAlgo      & 1.3\% & 1.4\% & 0 & 2 & 28 & 1372 & 1484 & -90.2\%  & 0.362 \\
IOUtracker  & 4.7\% & 4.4\% & 0 & 0 & 30 & 1526 & 1432 & -97.6\%  & 0.247 \\
SAM2+MOTRv2 & 90.8\% & 27.4\% & 21 & 7 & 2 & 3614 & 138  & -153.8\% & 0.141 \\
\bottomrule
\end{tabular}
\end{subtable}

\end{table}
\vspace{-0.7em}

\end{document}